\documentclass[runningheads,a4paper]{llncs}

\usepackage{amssymb}
\usepackage{graphicx}

\usepackage{sidecap}
\usepackage{lipsum,adjustbox}
\usepackage{tikz}
\usepackage{caption, graphicx}
\usepackage{float, amsmath, url}
\usepackage{hyperref, geometry}
\usetikzlibrary{positioning, fit, arrows.meta, shapes, arrows}
\usepackage{enumitem}

\usepackage{url}

\newcommand{\keywords}[1]{\par\addvspace\baselineskip
\noindent\keywordname\enspace\ignorespaces#1}

\newcommand{\etal}{\textit{et al.}}

\begin{document}

\mainmatter  


\title{Analysis of Convolutional Decoder for Image caption Generation}

\titlerunning{Convolutional Decoder for Image Caption Generation}

%
%

\author{Sulabh Katiyar%
\and Samir Kumar Borgohain}
\authorrunning{Sulabh Katiyar \& Samir Kumar Borgohain}


\institute{Department of Computer Science and Engineering\\
National Institute of Technology, Silchar, India \\}

%
%

\maketitle

\begin{abstract}
Recently Convolutional Neural Networks have been proposed for Sequence Modelling tasks such as Image Caption Generation. However, unlike Recurrent Neural Networks, the performance of Convolutional Neural Networks as Decoders for Image Caption Generation has not been extensively studied. In this work, we analyse various aspects of Convolutional Neural Network based Decoders such as Network complexity and depth, use of Data Augmentation, Attention mechanism, length of sentences used during training, etc on performance of the model. We perform experiments using Flickr8k and Flickr30k image captioning datasets and observe that unlike Recurrent Neural Network based Decoder, Convolutional Decoder for Image Captioning does not generally benefit from increase in network depth, in the form of stacked Convolutional Layers, and also the use of Data Augmentation techniques. In addition, use of Attention mechanism also provides limited performance gains with Convolutional Decoder. Furthermore, we observe that Convolutional Decoders show performance comparable with Recurrent Decoders only when trained using sentences of smaller length which contain up to 15 words but they have limitations when trained using higher sentence lengths which suggests that Convolutional Decoders may not be able to model long-term dependencies efficiently. In addition, the Convolutional Decoder usually performs poorly on CIDEr evaluation metric as compared to Recurrent Decoder.

\end{abstract}

\section{Introduction}

Image Caption Generation has witnessed tremendous work in the past decade where the goal is to generate a single sentence description of an image such that the generated sentence is similar to the way human beings generally describe images. When provided with an image, we have natural ability to extract salient information from the image and describe the image with a natural language sentence \cite{fei_fei}. However, for an automated system to be able to do this, it should be able to first represent the visual information in such a way that it can be easily processed and then it should be able to identify the relevant portions which contain semantic information about important objects and their relationships. In some cases, some information may be implicitly available in the scene but the salient objects may not be present. For example, given an image of a sports stadium filled with people, human beings are able to recognize that a game is being played and in some cases even recognize which game is being played. Similarly, the presence of a certain type of automobile (such as a Double Decker bus) may prompt a human being to recognize the probable location of the scene. However, such trivial understanding about the world that human beings possess through experience and are then able to transfer that learned information to a different task is very difficult for an automated system to learn.

Due to rapid progress in Deep Learning techniques, a lot of methods have been proposed for Image Caption Generation. For example, the Encoder-Decoder (\cite{sutskever}, \cite{cho}) and Attention (\cite{bahdanau}) based frameworks proposed for Machine Translation have been extended to Image Caption Generation as well, as proposed in Vinyals \etal \cite{vinyals}, Karpathy \etal \cite{karpathy}, Mao \etal \cite{mao} and Xu \etal \cite{attendtell}, where, instead of using the source sentence as input which is done in Machine Translation, the source image is used as input in Image Caption Generation. The Encoder module in Image Caption Generation is a Convolutional Neural Network which has been pre-trained using ImageNet dataset \cite{imagenet} on Object Detection task \cite{imagenetchallenge}. The output of last Convolutional Layer or the last fully connected layer is used as fixed length vectorized representation of the input image which is then used by the decoder module, which is generally a Recurrent Neural Network (RNN) \cite{elmanrnn} variant such as a Long Short Term Memory (LSTM) Network \cite{lstm}. Recently, methods proposed for Machine Translation using a Convolutional Neural Network as both Encoder and Decoder \cite{gehring} have been extended to Image Caption Generation  as well\cite{aneja}. 

Our work is focused on Image caption generation using a Convolutional Decoder. Recently, Aneja \etal \cite{aneja} have proposed an Image Caption Generation method using Convolutional Neural Network as decoder. However, unlike Caption Generation using Recurrent Neural Networks which has been studied in detail, since vast majority of Caption Generation methods use Recurrent Neural Networks, many aspects of Caption Generation using Convolutional Decoder have not been analysed such as the depth of Convolutional Decoder network, use of Data Augmentation techniques, use of Attention based guidance, length of sentences used for training, etc. 
To the best of our knowledge there does not exist a study which comparatively analyses such aspects of Convolutional Decoder based Image Captioning. 

The core contributions of our work are as follows:
\begin{itemize}
    \item We propose Image Caption Generation method using Convolutional Neural Network as decoder and compare the performance using different number of stacked Convolutional layers, Data Augmentation techniques, Attention Mechanism, variation in length of sentences during training, etc.
    \item We perform a detailed analysis of the performance of Image Caption Generation using Convolutional Decoder using two widely used datasets: Flickr8k\cite{flickr8k}, Flickr30k\cite{flickr30k}.
    \item We have observed that model does not show gains in performance when depth of network, in terms of number of stacked Convolutional Layers, is increased. Furthermore, Data Augmentation techniques don't offer improvement in performance with respect to commonly used metrics. This is unlike Recurrent Decoders where increase in the depth of network and the use of Data Augmentation techniques leads to increase in performance \cite{katiyar} \cite{bilstm}. Furthermore, attention mechanism leads to considerable performance gains for Recurrent Decoder, as demonstrated in Xu \etal \cite{attendtell}, but we observe that Convolutional Decoder achieves much lower performance gain when using Visual Attention. 
    \item In previous work of Aneja \etal \cite{aneja}, maximum length of sentences used for training was set to 15 words. Sentence larger than 15 words in length are removed during training and randomly replaced by those caption sentences of the image which have length of less than 15 words. In our work, we analyse the effect of maximum sentence length on model performance. To compare this aspect of Convolutional Decoder with Recurrent Decoder, we implement Encoder-Decoder based Image Captioning method using Recurrent Decoder as well. We observe that performance of Convolutional Decoder based Image Captioning model decreases drastically when maximum allowed length of sentences is more than 15 words unlike the Recurrent Decoder based Image Captioning model which has better performance when sentences with higher word length are allowed.
\end{itemize}

\section{Related Work}
\label{related_word}
The methods proposed for Image Caption Generation can be grouped into three broad categories: Retrieval based, Template based, Deep Learning based methods. Here we briefly discuss the three categories. The \textit{Retrieval Based} methods approach the task as a sentence retrieval problem where, given an image, a sentence is selected from a pre-defined pool of sentences and assigned as caption for the target image. In these methods, a collection of sentences is prepared which can reasonably describe most images in an environment. For example, Farhadi \etal \cite{farhadi} construct a meaning space for each target image and then compute the distance of each sentence in the pre-defined pool from this meaning space using a similarity metric. The meaning space is constructed as triplets of (object, action, scene).  In Ordonez \etal \cite{ordonez}, a large corpus of annotated images is constructed and semantic contents are extracted from the images. For a given target image, the image from dataset with most similar semantic content is found and it's caption is assigned to the target image as caption. In Mason \etal \cite{mason}, a collection of images is defined and given a target image, a visual similarity measure is used to find a set of images from the collection of all images which are similar to the target image. Then the captions of the retrieved images are used to calculate a word probability density score conditioned on the target image. Then the captions of the retrieved images are ranked using the word probability density score calculated in the last step. The sentence with the highest ranking is used as caption for the target image.

The \textit{Retrieval based} methods lead to captions which are grammatically correct since the collections of annotated images is prepared manually but are also unable to account for the diversity in semantic information in the target images since a lot of combinations of objects, their attributes and their relationships with each other can exist among different images which have similar scene settings. These methods also suffer from poor scalability because to account for a new scene setting, a large number of annotated images need to be collected to account for all possible combinations of the objects. Hence, some methods were proposed which attempted to generate novel descriptions for images using a set of hard-coded grammar and semantic rules. These are called \textit{Template Based} methods. For example, Kulkarni \etal \cite{kulkarni} recognize semantic contents of the target image, for which the caption is to be generated, using Conditional Random Fields and then plot the objects using a graph. The pairwise relationships between the objects in the graph are identified using statistical word collocation information. Li \etal \cite{li} extract visual information from the images and classify it in the form of objects, relationships and attributes. Then triplets of the form- [(adjective1,object1),preposition,(adjective2,object2)] are used to encode this information. Then the n-gram frequency statistics are calculated using word collocation information. Finally, using dynamic programming, the most likely combination of phrases is calculated and phrase fusion leads to the generated caption. These \textit{Template based} methods are able to generate more diverse captions but they also suffer from lack of scalability of the methods because a lot of templates need to be manually defined to account for various grammatical and semantic rules.

In order to train the model end-to-end and alleviate the issues related to manual creation of rules and similarity metrics, Deep Learning Methods have been proposed. This has mostly been possible due to tremendous progress in allied fields such as Machine Translation and Object Recognition. For Image Captioning, a fixed length representation of image is usually computed using a pre-trained Convolutional Neural Network (CNN) which was trained on the ImageNet dataset \cite{imagenet} for solving the Object Recognition problem \cite{imagenetchallenge}. The final classification layer of the CNN is discarded and a single vector is extracted from the final fully connected layer or a set of vectors is extracted from a later Convolutional Layer. Most Methods follow the Encoder-Decoder format originally proposed for Machine translation \cite{sutskever}, \cite{cho}. For example, Mao \etal \cite{mao} use a Recurrent Neural Network \cite{elmanrnn} to generate textual information conditioned on the previously generated words in the caption sentence and merge this information with the visual information which is then used to predict the next word. Karpathy \etal \cite{karpathy} use a Bidirectional Recurrent Neural Network to generate captions. Vinyals \etal \cite{vinyals} use the visual information as hidden and cell state initialization for the Long Short Term Memory (LSTM) \cite{lstm} Network and then generate the next word of the caption using the previously generated words. Donahue \etal \cite{donahue} merge the visual and textual information at each time-step and provide that as input to the LSTM decoder. Attention mechanism which was proposed for machine Translation in Bahdanau \etal \cite{bahdanau}, where the decoder learns to focus on input at each time step, has also been adapted for Image captioning in Xu at. al. \cite{attendtell}. More recently, Convolutional Neural Networks have been proposed for Machine Translation in Gehring \etal \cite{gehring} which have also been adapted for Image Caption Generation in Aneja \etal \cite{aneja}. 

Our work is similar to the work of Aneja \etal \cite{aneja} where the authors have proposed a Convolutional Neural Network as decoder. The authors have used a three layered CNN as decoder with 512 hidden units and have trained both the encoder and decoder CNNs, i.e., they fine tune the pre-trained CNN encoder which is used for extracting image features. However, most relevant methods proposed in the literature cite the results after training the decoder only and do not fine-tune the pre-trained encoder CNN. Hence, we also follow this approach to perform fair comparison with previous works. In addition, we also experiment with different number of layers, data augmentation, use of attention mechanism, length of sentences used for training the model, etc. and analyse the results. Furthermore, in Aneja \etal \cite{aneja}, the experiments have been performed using MSCOCO \cite{coco} dataset whereas we use Flickr8k \cite{flickr8k} and Flickr30k \cite{flickr30k} datasets.

\section{Convolutional Decoder for Image Caption Generation}
\label{cnn_decoder}
In this work we experiment with Convolutional Neural Network based decoder for Image Caption Generation. Unlike, the LSTM based decoders used in previous works \cite{vinyals}, \cite{karpathy}, \cite{mao}, \cite{attendtell}, \cite{donahue}, the Convolutional Decoder does not process the sentence sequentially and does not maintain a hidden or cell state which 'remembers' all relevant information it has previously encountered. Instead it processes the whole sequence parallelly and models the sequential information using 1-D Convolutions. 

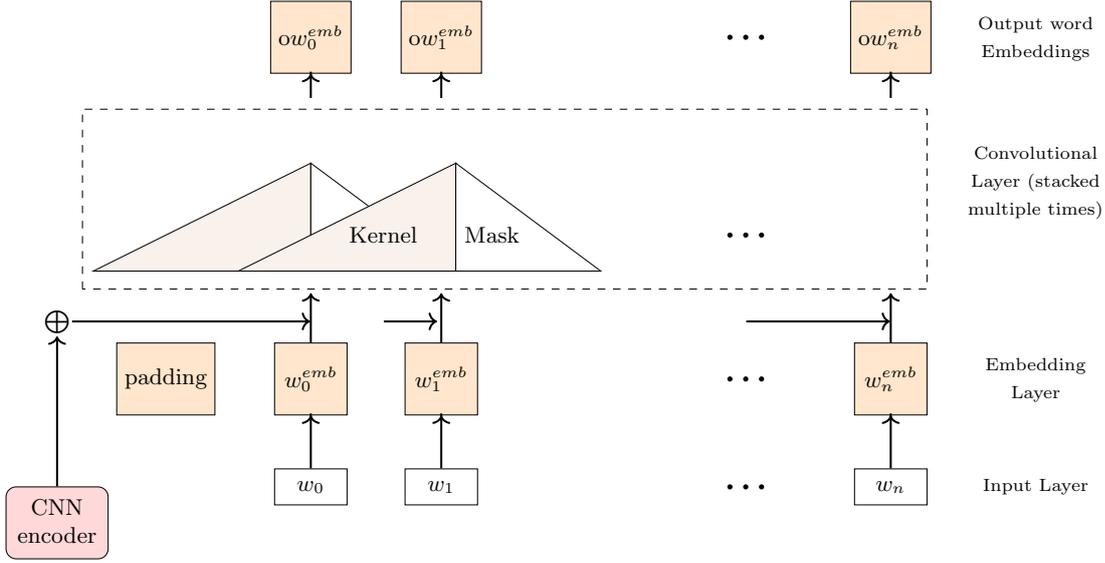
\begin{figure}
\tikzstyle{startstop} = [rectangle, rounded corners, minimum width=1cm, minimum height=1cm,text centered, draw=black]
\tikzstyle{io} = [trapezium, trapezium left angle=70, trapezium right angle=110, minimum width=1cm, minimum height=.5cm, text centered, draw=black]
\tikzstyle{input} = [rectangle, minimum width=1.0cm, minimum height=0.5cm, text centered, draw=black]
\tikzstyle{LSTM} = [rectangle, rounded corners, minimum width=1cm, minimum height=3cm, text centered, draw=black, fill=blue!20]
\tikzstyle{SOFT} = [rectangle, rounded corners, minimum width=2cm, minimum height=1cm, text centered, draw=black, fill=blue!10]
\tikzstyle{decision} = [diamond, minimum width=1cm, minimum height=.2cm, text centered, draw=black]
\tikzstyle{fcimagestyle} = [rectangle, rounded corners, minimum width=1cm, minimum height=1cm,text centered, draw=black, fill=red!30]
\tikzstyle{fctextstyle} = [rectangle, rounded corners, minimum width=1cm, minimum height=1cm,text centered, draw=black, fill=orange!20]
\tikzstyle{embedstyle} = [rectangle, sharp corners, minimum width=1cm, minimum height=1cm,text centered, draw=black, fill=orange!20]
\tikzstyle{NASNET} = [rectangle, rounded corners, minimum width=1cm, minimum height=1.0cm,text centered, draw=black, fill=red!15]
\tikzstyle{dotbox} = [rectangle, rounded corners, minimum width=1cm, minimum height=1cm,text centered, draw=none]
\tikzstyle{atimagestyle} = [rectangle, rounded corners, minimum width=1cm, minimum height=1cm,text centered, draw=black, fill=brown!30]

\tikzstyle{arrow} = [thick,->]
\centering
\resizebox{1.0\textwidth}{!}{%
\begin{tikzpicture}[node distance = 1.0cm]

    \node (word0) [input]{$w_0$};
    \node (word1) [input, right of= word0, node distance = 1.8cm]{$w_1$};
    \node (dotword) [dotbox, right of= word0, node distance = 6cm] {\Huge{...}};
    \node (wordn) [input, right of= word0, node distance = 8cm]{$w_n$};
    \node (wdesc) [dotbox, right of= wordn, node distance = 2cm] {\scriptsize Input Layer};
    \node (emb0) [embedstyle,above of=word0,node distance=1.5cm] {$w_0^{emb}$ };
    \node (embminus) [embedstyle,left of=emb0,node distance=2.0cm] {padding};
    \node (emb1) [embedstyle,above of=word1,node distance=1.5cm] {$w_1^{emb}$ };
    \node (dotemb) [dotbox, right of= emb0, node distance=6cm] {\Huge{...}};
    \node (embn) [embedstyle,above of=wordn,node distance=1.5cm] {$w_n^{emb}$ };
    \node (edesc) [dotbox, right of= embn, node distance = 2cm] {\begin{tabular}{c}
         \scriptsize Embedding  \\
         \scriptsize Layer
    \end{tabular}};
    \draw[fill=brown!10]  (0, 3)-- (0,4.5)-- (-3.0,3)-- (0, 3);
    \draw[fill=white]  (0, 3)-- (0,4.5)-- (2.0,3)-- (0, 3);
    \draw[fill=brown!10]  (2, 3)-- (2,4.5)-- (-1.0,3)-- (2, 3);
    \draw[fill=white]  (2, 3)-- (2,4.5)-- (4.0,3)-- (2, 3);
    \node (dotkernel) [dotbox,above of=emb0, node distance=2cm, xshift=1cm] {Kernel};
    \node (dotmask) [dotbox,above of=emb0, node distance=2cm, xshift=2.5cm] {Mask};
    \node (dotconv) [dotbox, right of= dotmask, node distance=3.5cm] {\Huge{...}};
    
    \node (plus1) [dotbox, left of= embminus, node distance=1.5cm, yshift=0.8cm] {$\bigoplus$};
    \node (plusarrow) [dotbox, right of= plus1, node distance=4.0cm, yshift=0.0cm] {};
    
    \node (cnn) [NASNET, left of= embminus, node distance=2cm, xshift = 0.5cm, yshift=-2cm] {\begin{tabular}{c}
         CNN \\
         encoder
    \end{tabular}};

    \draw [arrow, ->] ([xshift = 0.0cm]cnn.north) -- ([yshift = 0.3cm]plus1.south);

    \draw [arrow, ->] (word0.north) |- (emb0.south);
    \draw [arrow, ->] (emb0.north) -- ([xshift = -1cm, yshift = -0.3cm]dotkernel.south);
    \draw [arrow, ->] (word1.north) |- (emb1.south);
    \draw [arrow, ->] (emb1.north) -- ([xshift = 0.8cm, yshift = -0.3cm]dotkernel.south);
    \draw [arrow, ->] (wordn.north) |- (embn.south);
    \draw [arrow, ->] (embn.north) -- ([xshift = 7cm, yshift = -0.3cm]dotkernel.south);
    \draw [arrow, ->] ([xshift = -0.3cm]plus1.east) -- (plusarrow.west);
    \draw [arrow, ->] ([xshift = 4.0cm]plus1.east) -- ([xshift = 1.75cm]plusarrow.west);
    \draw [arrow, ->] ([xshift = 9.0cm]plus1.east) -- ([xshift = 8.0cm]plusarrow.west);
    
    \draw[dashed]  (-3.15, 2.75)-- (-3.15,5.25)-- (8.5,5.25)-- (8.5,2.75)--(-3.15, 2.75);
    
    \node (oemb0) [embedstyle,above of=emb0,node distance=4.75cm] {o$w_0^{emb}$ };
    \node (oemb1) [embedstyle,above of=emb1,node distance=4.75cm] {o$w_1^{emb}$ };
    \node (oembn) [embedstyle,above of=embn,node distance=4.75cm] {o$w_n^{emb}$ };
    \node (dotoemb) [dotbox, right of= oemb0, node distance = 6cm] {\Huge{...}};
    \node (odesc) [dotbox, right of= oemb0, node distance = 10cm] {\begin{tabular}{c}
    \scriptsize Output word  \\
    \scriptsize Embeddings
    \end{tabular}};
    \node (cdesc) [dotbox, right of= oemb0, node distance = 10cm, yshift = -2.0cm] {\begin{tabular}{c}
    \scriptsize Convolutional   \\
    \scriptsize Layer (stacked\\
    \scriptsize multiple times)
    \end{tabular}};
    \node (outputarrow) [dotbox, below of= oemb0, node distance=1.0cm, yshift=0.0cm] {};
    \draw [arrow, ->] ([yshift = -0.35cm]outputarrow.north) |- (oemb0.south);
    \draw [arrow, ->] ([xshift = 1.8cm,yshift = -0.35cm]outputarrow.north) |- (oemb1.south);
    \draw [arrow, ->] ([xshift = 8cm,yshift = -0.35cm]outputarrow.north) |- (oembn.south);

\end{tikzpicture}
}

\centering
\caption{An overview of the Image Caption Generation using a Convolutional Decoder (CNN+CNN method).  \\ }
\label{cnn_decoder_diagram}
\end{figure}

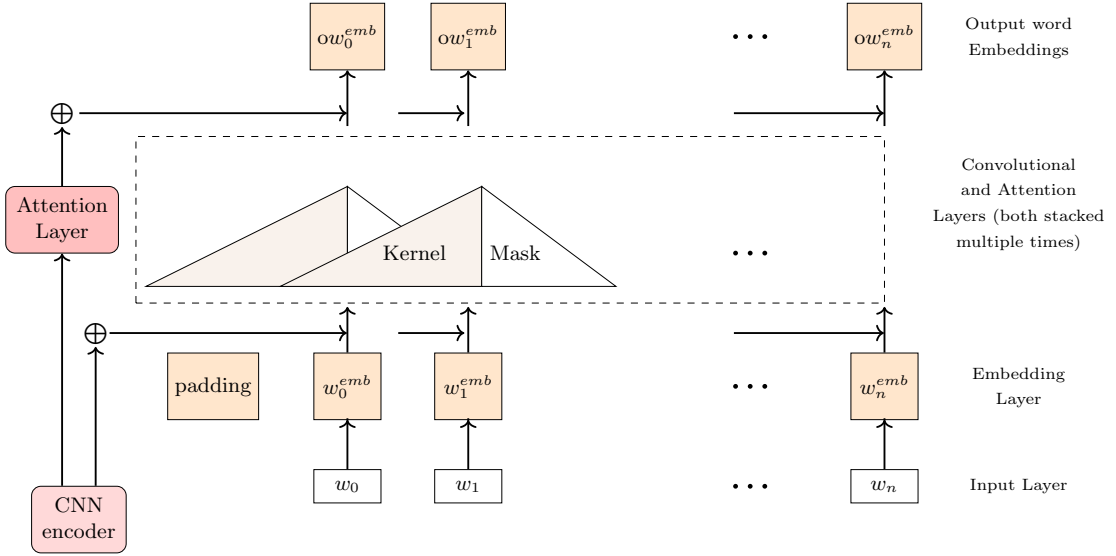
\begin{figure}
\tikzstyle{startstop} = [rectangle, rounded corners, minimum width=1cm, minimum height=1cm,text centered, draw=black]
\tikzstyle{io} = [trapezium, trapezium left angle=70, trapezium right angle=110, minimum width=1cm, minimum height=.5cm, text centered, draw=black]
\tikzstyle{input} = [rectangle, minimum width=1.0cm, minimum height=0.5cm, text centered, draw=black]
\tikzstyle{LSTM} = [rectangle, rounded corners, minimum width=1cm, minimum height=3cm, text centered, draw=black, fill=blue!20]
\tikzstyle{SOFT} = [rectangle, rounded corners, minimum width=2cm, minimum height=1cm, text centered, draw=black, fill=blue!10]
\tikzstyle{decision} = [diamond, minimum width=1cm, minimum height=.2cm, text centered, draw=black]
\tikzstyle{fcimagestyle} = [rectangle, rounded corners, minimum width=1cm, minimum height=1cm,text centered, draw=black, fill=red!30]
\tikzstyle{fctextstyle} = [rectangle, rounded corners, minimum width=1cm, minimum height=1cm,text centered, draw=black, fill=orange!20]
\tikzstyle{embedstyle} = [rectangle, sharp corners, minimum width=1cm, minimum height=1cm,text centered, draw=black, fill=orange!20]
\tikzstyle{NASNET} = [rectangle, rounded corners, minimum width=1cm, minimum height=1.0cm,text centered, draw=black, fill=red!15]
\tikzstyle{dotbox} = [rectangle, rounded corners, minimum width=1cm, minimum height=1cm,text centered, draw=none]
\tikzstyle{atimagestyle} = [rectangle, rounded corners, minimum width=1cm, minimum height=1cm,text centered, draw=black, fill=red!25]

\tikzstyle{arrow} = [thick,->]
\centering
\resizebox{1.0\textwidth}{!}{%
\begin{tikzpicture}[node distance = 1.0cm]

    \node (word0) [input]{$w_0$};
    \node (word1) [input, right of= word0, node distance = 1.8cm]{$w_1$};
    \node (dotword) [dotbox, right of= word0, node distance = 6cm] {\Huge{...}};
    \node (wordn) [input, right of= word0, node distance = 8cm]{$w_n$};
    \node (wdesc) [dotbox, right of= wordn, node distance = 2cm] {\scriptsize Input Layer};
    \node (emb0) [embedstyle,above of=word0,node distance=1.5cm] {$w_0^{emb}$ };
    \node (embminus) [embedstyle,left of=emb0,node distance=2.0cm] {padding};
    \node (emb1) [embedstyle,above of=word1,node distance=1.5cm] {$w_1^{emb}$ };
    \node (dotemb) [dotbox, right of= emb0, node distance=6cm] {\Huge{...}};
    \node (embn) [embedstyle,above of=wordn,node distance=1.5cm] {$w_n^{emb}$ };
    \node (edesc) [dotbox, right of= embn, node distance = 2cm] {\begin{tabular}{c}
         \scriptsize Embedding  \\
         \scriptsize Layer
    \end{tabular}};
    \draw[fill=brown!10]  (0, 3)-- (0,4.5)-- (-3.0,3)-- (0, 3);
    \draw[fill=white]  (0, 3)-- (0,4.5)-- (2.0,3)-- (0, 3);
    \draw[fill=brown!10]  (2, 3)-- (2,4.5)-- (-1.0,3)-- (2, 3);
    \draw[fill=white]  (2, 3)-- (2,4.5)-- (4.0,3)-- (2, 3);
    \node (dotkernel) [dotbox,above of=emb0, node distance=2cm, xshift=1cm] {Kernel};
    \node (dotmask) [dotbox,above of=emb0, node distance=2cm, xshift=2.5cm] {Mask};
    \node (dotconv) [dotbox, right of= dotmask, node distance=3.5cm] {\Huge{...}};
    
    \node (plus1) [dotbox, left of= embminus, node distance=1.5cm, yshift=0.8cm, xshift = -0.25cm] {$\bigoplus$};
    \node (plusarrow) [dotbox, right of= plus1, node distance=4.25cm, yshift=0.0cm] {};
    
    \node (cnn) [NASNET, left of= embminus, node distance=2cm, xshift = -0.0cm, yshift=-2cm] {\begin{tabular}{c}
         CNN \\
         encoder
    \end{tabular}};
    \node (att) [atimagestyle, above of= cnn, node distance=4.5cm, xshift = -0.25cm] {\begin{tabular}{c}
        Attention \\
        Layer
    \end{tabular}};
    \node (plus2) [dotbox, above of= att, node distance=1.6cm, xshift=0.0cm] {$\bigoplus$};
    \draw [arrow, ->] ([xshift = 0.0cm]att.north) -- ([yshift = 0.3cm]plus2.south);
    
    \draw [arrow, ->] ([xshift = 0.25cm]cnn.north) -- ([yshift = 0.3cm]plus1.south);
    \draw [arrow, ->] ([xshift = -0.25cm]cnn.north) -- ([xshift = -0.0cm]att.south);

    \draw [arrow, ->] (word0.north) |- (emb0.south);
    \draw [arrow, ->] (emb0.north) -- ([xshift = -1cm, yshift = -0.3cm]dotkernel.south);
    \draw [arrow, ->] (word1.north) |- (emb1.south);
    \draw [arrow, ->] (emb1.north) -- ([xshift = 0.8cm, yshift = -0.3cm]dotkernel.south);
    \draw [arrow, ->] (wordn.north) |- (embn.south);
    \draw [arrow, ->] (embn.north) -- ([xshift = 7cm, yshift = -0.3cm]dotkernel.south);
    \draw [arrow, ->] ([xshift = -0.3cm]plus1.east) -- (plusarrow.west);
    \draw [arrow, ->] ([xshift = 4.0cm]plus1.east) -- ([xshift = 1.75cm]plusarrow.west);
    \draw [arrow, ->] ([xshift = 9.0cm]plus1.east) -- ([xshift = 8.0cm]plusarrow.west);
    
    \draw[dashed]  (-3.15, 2.75)-- (-3.15,5.25)-- (8,5.25)-- (8,2.75)--(-3.15, 2.75);
    
    \node (oemb0) [embedstyle,above of=emb0,node distance=5.25cm] {o$w_0^{emb}$ };
    \node (oemb1) [embedstyle,above of=emb1,node distance=5.25cm] {o$w_1^{emb}$ };
    \node (oembn) [embedstyle,above of=embn,node distance=5.25cm] {o$w_n^{emb}$ };
    \node (dotoemb) [dotbox, right of= oemb0, node distance = 6cm] {\Huge{...}};
    \node (odesc) [dotbox, right of= oemb0, node distance = 10cm] {\begin{tabular}{c}
    \scriptsize Output word  \\
    \scriptsize Embeddings
    \end{tabular}};
    \node (cdesc) [dotbox, right of= oemb0, node distance = 10cm, yshift = -2.5cm] {\begin{tabular}{c}
    \scriptsize Convolutional   \\
    \scriptsize and Attention \\
    \scriptsize Layers (both stacked \\
    \scriptsize multiple times)
    \end{tabular}};
    \node (plusarrow2) [dotbox, right of= plus2, node distance=4.25cm, yshift= -0.1cm]{};
    \draw [arrow, ->] ([yshift = -0.6cm]plusarrow2.north) -- (oemb0.south);
    \draw [arrow, ->] ([yshift = -0.6cm, xshift = 1.8cm]plusarrow2.north) -- (oemb1.south);
    \draw [arrow, ->] ([yshift = -0.6cm, xshift = 8cm]plusarrow2.north) -- (oembn.south);
    \draw [arrow, ->] ([yshift = 0.0cm, xshift = -0.25cm]plus2.east) -- ([yshift = 0.1cm, xshift = 0.5cm]plusarrow2.west);
    \draw [arrow, ->] ([yshift = 0.0cm, xshift = 4.5cm]plus2.east) -- ([yshift = 0.1cm, xshift = 2.25cm]plusarrow2.west);
    \draw [arrow, ->] ([yshift = 0.0cm, xshift = 9.5cm]plus2.east) -- ([yshift = 0.1cm, xshift = 8.5cm]plusarrow2.west);
    
\end{tikzpicture}
}

\centering
\caption{An overview of the Image Caption Generation using a Convolutional Decoder with Attention Mechanism (CNN+CNN+Attention method).  \\ }
\label{cnn_decoder_attention_diagram}
\end{figure}

We describe The Encoder-Decoder based Image Captioning using Convolutional Decoder in Figure \ref{cnn_decoder_diagram} and the Attention based Convolutional Decoder method in Figure \ref{cnn_decoder_attention_diagram}. We refer to both the methods as \textit{CNN+CNN} method (Convolutional Encoder and Convolutional Decoder) and \textit{CNN+CNN+Att} method (Convolutional Encoder and Convolutional Decoder with Attention), respectively.

Given an image $I$, the Image Caption Generation task is to generate a sequence of words $S = \{x_1, x_2,x_3, ..., x_n \}$. Since, the generation of words depends on previous words of the sentence, we can model the caption generation as probabilistic problem such that probability of a word $x_i$ can be represented as, 
\begin{equation}
    p_{i, \Omega}(x_i \arrowvert I) = p_{i, \Omega}(x_i \arrowvert x_{1:i-1}, I)
\end{equation}
where $\Omega$ is the set of all parameters of the model and $\{ x_1, x_2, x_3, ..., x_{i-1}\}$ are the words in the sentence occurring before the word $x_i$. During training, the aim is to maximize the log likelihood of the generated sentence, i.e., to maximize,
\begin{equation*}
        arg \max_\Omega \sum_i log p(x_{1:L_{I}}|I, \Omega), \hspace{2cm} i \in (1,2,3,...,L_I)
\end{equation*}
where, $L_I$ is the length of caption sentence for Image $I$. Since, words in a sentence depend on the previous words, we can use chain rule to compute likelihood of the caption sentence as,
\begin{equation}
       log p(x_{1:L_I}|I, \Omega) = log p(x_1|I, \Omega) + \sum_{j=2}^{L_I} log p(x_j|I, x_{1:j-1}, \Omega)
\end{equation}
where $x_1$ is the first word in the sentence and hence its generation depends only on the input image. All possible words that can be generated comprise the vocabulary of the model, i.e., $x_i \in \mathcal{X}, i \in L_I$ where $\mathcal{X}$ is the vocabulary of the model. We remove words which occur less than 5 times in the dataset so for Flickr8k dataset \cite{flickr8k} $\arrowvert \mathcal{X} \arrowvert = 2362$ and for Flickr30k dataset \cite{flickr30k}, $\arrowvert \mathcal{X} \arrowvert = 7002$. The vocabulary also includes the special 'start' and 'end' tokens which signal the beginning and end of the generation process and also the special $<PAD>$ token for padding the end of sentence which is shorter than the maximum allowed length of sentence.

In both the methods, the words are encoded as dense word vectors which are learned as weights of the Word Embedding layer which learns a function to map one-hot word vectors into low dimensional dense vectors, as \begin{equation}
    x_i = f_{embedding}(x_{i}^{one-hot})
\end{equation}
where $i \in (1,2,3,..., \mathcal{X})$,  $x_i \in \mathrm{R}^{\mathrm{E}}$ where $\mathrm{E}$ is the dimension of Word Embeddings and $x_{i}^{one-hot} \in \mathrm{R}^{\mathcal{X}}$. The word vector representations (word embeddings) are then concatenated with image features extracted from pre-trained CNN Encoder and passed to the CNN decoder layer. Unlike the method proposed by Aneja \etal \cite{aneja} where the CN encoder is also trained along with the CNN decoder, we do not train the CNN encoder because its consistent with approaches used in other methods proposed in the literature such as Vinyals \etal \cite{vinyals}, Mao \etal \cite{mao}, Karpathy \etal \cite{karpathy}, Vinyals \etal \cite{attendtell} and thus allows a fair comparison with previously proposed methods. 
The Convolutional Layer performs masked convolutions over the concatenated input word embedding and image features. We have experimented with different number of Convolutional layers in our work (from one layer to four layers) as opposed to three layers used in Aneja \etal \cite{aneja}. The activation layer is composed of Gated Linear Units (GLU) and the the width of receptive fields of masked convolutions is five. The Convolutional Layer outputs 512-dimensional vector for each word. We also experiment with different values of maximum length of sentences used for training. The output of last Convolutional Layer is passed through a Softmax layer which converts the output into word probability vector of length $\mathcal{X}$ comprising relative probability of occurrence of each word in the vocabulary. 
In \textit{CNN+CNN+Att} method, the output of Convolutional Layer is projected on the vector space comprising word embedding vectors for each word and then attended image feature vector is calculated for each word conditioned on the output of masked convolutions.

\section{Experiments}
\label{experiments}
In this section, we describe the details of our experiments. We analyse the effect of layer depth, use of Image Transforms for Data Augmentation, use of Visual Attention and different values of maximum lengths of sentence used for training on the performance of the model. For all the experiments, we have used ResNet-152 CNN \cite{resnet} pretrained on ImageNet Object Detection task \cite{imagenetchallenge} and we extract image features from the last convolutional layer of the network. It has been observed that use of ResNet CNN leads to increase in performance of Image Captioning models \cite{cnn_variants}. For inference using trained model, we have used Beam Width of 3. Since we observe that the best scores are obtained when the model is trained using only one Convolutional Layer for the \textit{CNN+CNN} model and two Convolutional layers for \textit{CNN+CNN+Att} model, we have quoted the evaluation results of captions generated using model trained with one Convolutional Layer for CNN+CNN model and two layers for \textit{CNN+CNN+Att} model, unless stated otherwise (as in Section \ref{decoder_depth}). We have performed our experiments using NVIDIA Quadro RTX 4000 Graphics processor which has around 7 GBs of graphics memory. All our experiments have been performed with a batch size of 10. In Section \ref{sentence_length}, we describe the experiments performed using an LSTM decoder to comparatively analyse the performance of Convolutional and Recurrent Decoder when using different values of maximum length of sentences used for training the models. The experiment using LSTM decoder are performed using a batch size of 32 and a beam width of 3 is used for inference.

\subsection{Compared Methods}
\label{compared_methods}

We compare the performance with relevant methods proposed in the literature. Vinyals \etal \cite{vinyals} use a Long Short Term Memory (LSTM) Network to generate captions which takes visual information as the initial hidden and cell states. Mao \etal \cite{mao} use a Recurrent Neural Network which takes as input previous word embedding vectors and the generated output is then merged with image feature vectors which have been mapped to the same vector space. The merged output is then passed through a prediction layer. Karpathy \etal \cite{karpathy} use a Bidirectional Recurrent Neural Network to generate the captions. Donahue \etal \cite{donahue} use an LSTM generator similar to the work of Vinyals \etal \cite{donahue} but they introduce visual information at each time-step in the decoder. Xu \etal \cite{attendtell} use an LSTM as decoder but also introduce attention mechanism where the set of image vector representations obtained for different image sub-regions are passed through attention layer which generates attended visual information containing a set of feature vectors from those image sub-regions which are important at the current time-step.  We discuss the results in Section \ref{comparison_with_other_methods}. BLEU, METEOR, CIDEr and ROUGE-L metrics are used for evaluation \cite{chen}.

\subsection{Number of stacked Convolutional Layers used in the model}
\label{number_of_layers}
In previous methods, it has been observed that increasing the depth of decoder by stacking multiple layers of RNN or LSTM decoder also leads to increase in performance of the model. In Wang \etal \cite{bilstm}, a multi-layer Bidirectional LSTM has been used for Image Captioning task. In Katiyar \etal \cite{katiyar}, a three-layer stacked Deep-LSTM model has been proposed and data augmentation is used to generate captions. We explore this aspect with respect to Convolutional Decoder for both Encoder-Decoder (\textit{CNN+CNN}) and Attention based (\textit{CNN+CNN+Att}) methods. Using Convolutional Decoder, Aneja \etal \cite{aneja} propose a method in which three convolutional layers are stacked over each other and the output of the lower layer is the input of the next layer. Residual connections are also used between successive layers. In our experiments, we observe that the best performance is achieved using one convolutional layer for the Encoder-Decoder based method (\textit{CNN+CNN}) and using two layers for Attention based method (\textit{CNN+CNN+Att}). We discuss the results in Section \ref{decoder_depth}. 

\subsection{Data Augmentation with Image Transforms}
\label{data_augmentation}
We have randomly applied Image Transforms to a section of images in the dataset such that the model encounters transformed versions of images in different epochs of training thereby leading to overall more images being 'seen' by the model during training. We have have used Horizontal and Vertical Flipping, Random Rotations to 90, 180 and 270 degrees and Perspective Transforms. 

In perspective transformations \cite{perspective_paper}, the reference plane of the image is changed so that image appears to be from a different 'perspective' without changing the overall spatial distribution of objects with respect to each other. In Figure \ref{perspective_transforms}, we provide some examples of flipping, rotation and perspective transforms that we have used for data augmentation in our experiments.

\begin{table}[h]
    \centering
    \begin{tabular}{ccccc}
         \includegraphics[height = 3cm, width = 3cm]{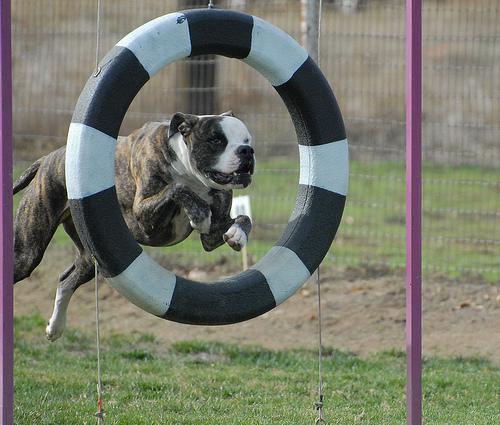}
         &\includegraphics[height = 3cm, width = 3cm]{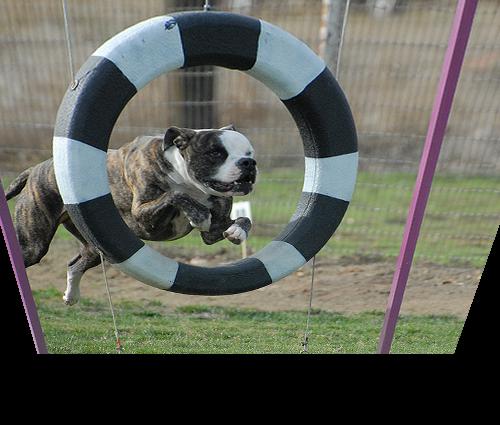}
         &\includegraphics[height = 3cm, width = 3cm]{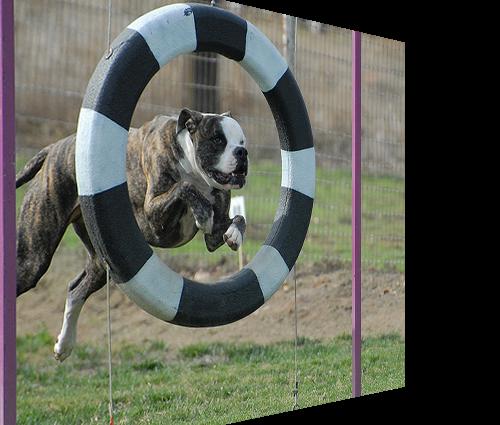}
         &\includegraphics[height = 3cm, width = 3cm]{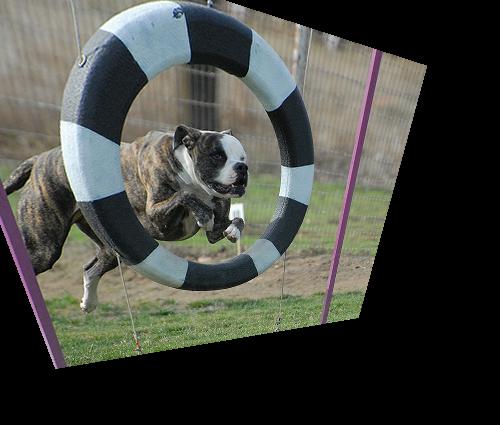}
         &\includegraphics[height = 3cm, width = 3cm]{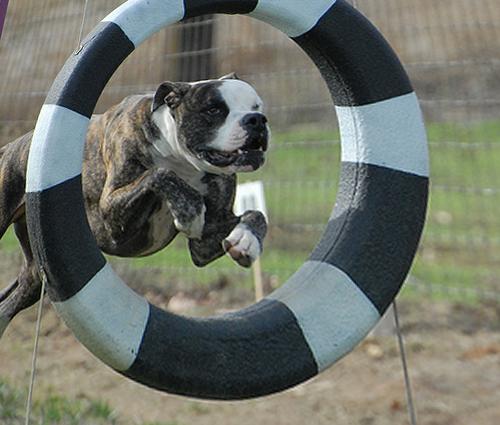}  \\
         (a)&(b)&(c)&(d)&(e) \\
         \includegraphics[height = 3cm, width = 3cm]{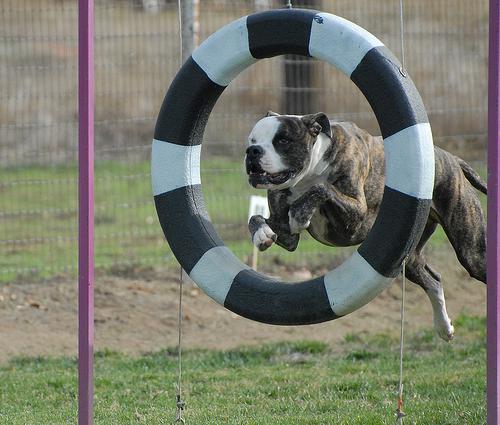}
         &\includegraphics[height = 3cm, width = 3cm]{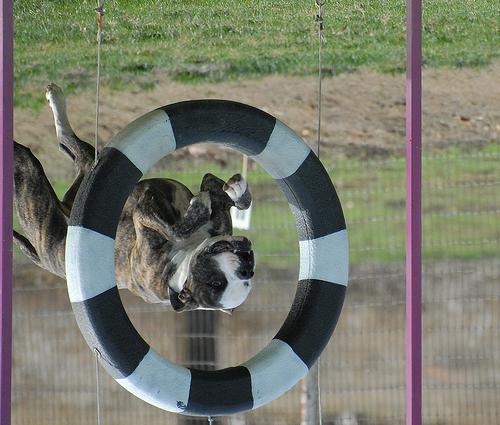}
         &\includegraphics[height = 3cm, width = 3cm]{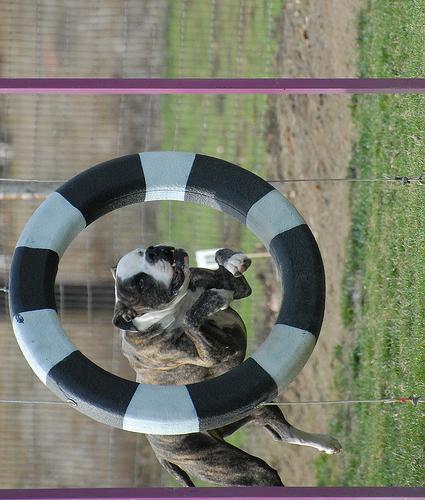}
         &\includegraphics[height = 3cm, width = 3cm]{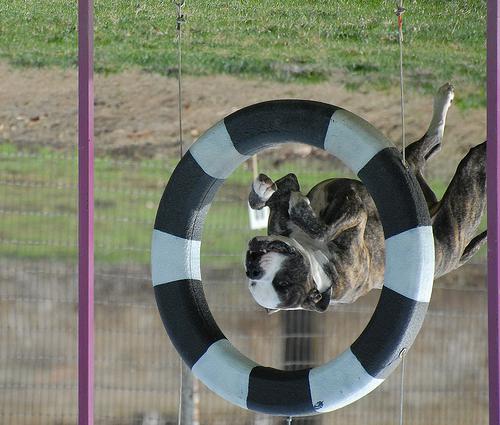}
         &\includegraphics[height = 3cm, width = 3cm]{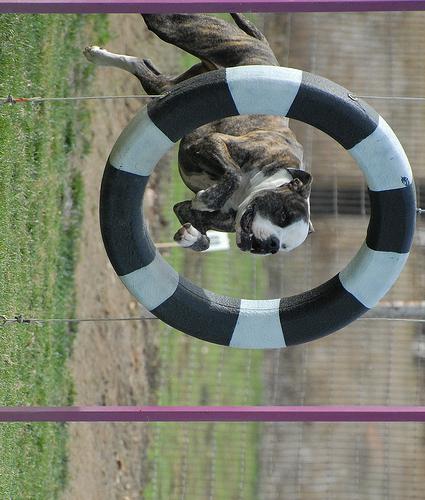} \\
         (f)&(g)&(h)&(i)&(j) \\
    \end{tabular}
    \captionof{figure}{Sub-Figure (a) is an image in Flickr8k dataset. Upon applying Perspective Transforms the transformed images are displayed in Sub-Figures (b)-(e). Images in Sub-Figures (f)-(j) are the transformed images after the transformations of Horizontal Flip, Vertical Flip, Rotation by 90 degrees, Rotation by 180 degrees, Rotation by 270 degrees, respectively, have been applied on Image in Sub-Figure(a).}
    \label{perspective_transforms}
\end{table}

In our experiments, we have used five types of transforms for Data Augmentation:
\begin{itemize}
    \item \textit{Random Horizontal}: The images are flipped horizontally about the y-axis, i.e., the vertical mirror image of the image is used. The images are flipped with a probability of 0.5 which means in half the number of epochs model 'sees' the original image and in other half it 'sees' flipped image.
    \item \textit{Random Vertical}: The images are flipped vertically, i.e., the mirror image about the x-axis is used with a probability of 0.5. 
    \item \textit{Random Flipping}: Images are flipped with probability of 0.5. And for each case of flipping transformation, either horizontal or vertical flipping is applied with a probability of 0.5. Hence, in images are flipped with an effective probability of 0.5 and horizontal and vertical transforms are applied with an effective probability of 0.25 each.
    \item \textit{Random Rotation}: Images are rotated to either 90, 180 or 270 degrees in their reference plane with a probability of 0.6 which is equally distributed among all the rotation transforms. Hence, the original image is provided to the model with a probability of 0.4 and each rotation is applied with a probability of 0.2.
    \item \textit{Random Perspective}: In random perspective transformation, the reference plane of the image is changed. In our experiments, perspective transforms are applied with a probability of 0.5. 
\end{itemize}

\subsection{Variation in sentence length}
\label{sentence_length}
We experiment with various values of maximum sentence length, i.e., the number of steps for masked convolutions, while selecting sentences in training data. In Aneja \etal \cite{aneja}, sentences greater than 15 words in length are discarded from the training data and replaced with other captions of the same image. We follow the same approach while selecting sentences for inclusion in training data. We observe that performance of the model varies widely when using different values of maximum sentence length. Specifically, the performance is better when using sentences of shorter length only but when larger length sentences are allowed then performance starts to drop.

In order to analyse the performance with respect to variation in maximum allowed values of sentence lengths, we have implemented Encoder-Decoder Image Captioning method using LSTM decoder. We call this method \textit{CNN+LSTM} method since a pre-trained CNN encoder is used for image features extraction and an LSTM decoder is used for sentence generation. This method is similar to the method proposed in Vinyals \etal \cite{vinyals} since we use the image features to generate initial hidden and cell states of the LSTM. There are some implementation differences:
\begin{itemize}
    \item We use ResNet CNN \cite{resnet} for feature extraction instead of Inception CNN used in Vinyals \etal \cite{vinyals}. In previous work it has been observed that using ResNet CNN allows better performance in Image captioning \cite{cnn_variants}. 
    \item We do not use batch normalization since we observe this does not lead to improvement in evaluation scores.
    \item We do not use model ensembles since most other methods proposed in previous work \cite{mao}, \cite{karpathy},  \cite{attendtell},  \cite{donahue},\cite{aneja} do not use model ensembles for training. 
\end{itemize}
We discuss the results in Section \ref{sentence_length}.

\section{Results and Discussion}
\label{results}
In this section we demonstrate the following results:
\begin{itemize}
    \item Convolutional Decoder based Image Captioning method provides results which are comparable with Captioning methods which use Recurrent Neural Network as decoder. However, attention based method using Convolutional Decoder does not provide tangible performance gains as compared to method using Convolutional Decoder without attention. 
    \item In previous work of Aneja \etal \cite{aneja}, three layered Convolutional Decoder has been used where separate attention mechanism is used at each layer. In our work, we observe that best performance is achieved while using a single layered Convolutional decoder for Encoder-decoder based approach and two layered Convolutional Decoder for Attention based approach. 
    \item Image Data Augmentation using Image Transforms have been shown to provide performance gains in Recurrent Decoder based Image Captioning methods \cite{katiyar}, \cite{bilstm}. However, Image Data Augmentation does not provide performance improvements in Convolutional Decoder based Captioning method.
    \item Convolutional Decoder based Image Captioning model shows better performance when trained using sentence length of up to 15 words. However, performance decreases significantly when it is trained using sentences of higher word lengths.
    \item On CIDEr evaluation metric, Recurrent Decoder outperforms Convolutional Decoder by around 8 to 16 points, depending on maximum length of the sentences used for training.
\end{itemize}

\subsection{Comparison with other methods in literature}
\label{comparison_with_other_methods} We compare the performance of \textit{CNN+CNN} and \textit{CNN+CNN+Att} methods on Flickr8k and Flickr30k datasets. As described in Section \ref{decoder_depth}, the best performance is generally achieved when using one Convolutional Layer and those are the results that we quote in this section. In Table \ref{comparison_sota}, we compare the results of Convolutional Decoder using Encoder-Decoder and Attention based frameworks on Flickr8k and Flickr30k datasets.

\begin{table}[h]
    \centering
    \begin{tabular}{p{2.75cm}|p{1.5cm}p{1.5cm}p{1.5cm}p{1.5cm}p{1.5cm}p{1.5cm}p{2cm}}
    \hline \noindent
         Method
         & BLEU-1
         & BLEU-2
         & BLEU-3
         & BLEU-4
         & METEOR
         & CIDEr
         & ROUGE-L \\ \hline 
          \multicolumn{8}{l}{Flickr8k Dataset} \\ 
          \hline      
          Karpathy \etal \cite{karpathy} & 0.579 & 0.383 & 0.245 & 0.160 & \_ & \_ & \_  \\
          Vinyals \etal \cite{vinyals} & 0.63 & 0.41 & 0.27 & \_& \_ & \_ & \_ \\
          Xu \etal \cite{attendtell} & 0.67 & 0.448 & 0.299 & 0.195 & 0.1893 & \_ & \_ \\
         
         \textit{CNN+CNN} & 0.6279 & 0.4439 & 0.3022 & 0.2038 & 0.1924 & 0.4748 & 0.4544\\
         \textit{CNN+CNN+Att} & 0.6291 & 0.4443 & 0.3032 & 0.2060 & 0.1944 & 0.4851 & 0.4556 \\ \hline
         
          \multicolumn{8}{l}{Flickr30k dataset} \\ \hline
          Mao \etal \cite{mao} & 0.60 & 0.41 & .028 & 0.19 & \_ & \_ & \_ \\
          Donahue \etal \cite{donahue} & 0.5872 & 0.3906 & 0.2512 & 0.1646 & \_ & \_ & \_ \\
          Karpathy \etal \cite{karpathy} & 0.573 & .369 & 0.240 & 0.157 & \_ & \_ & \_ \\
          Vinyals \etal \cite{vinyals} & 0.663 & 0.423 & 0.277 & 0.183 & \_ & \_ & \_  \\
          Xu \etal \cite{attendtell} & 0.667 & 0.434 & 0.288 & 0.191 & 0.1849 & \\
          \textit{CNN+CNN} & 0.6432 & 0.4495 & 0.3108 & 0.2123 & 0.1774 & 0.3831 & 0.4344 \\ 
          \textit{CNN+CNN+Att} & 0.6393 & 0.4428 & 0.3053 & 0.2088 & 0.1778 & 0.3810 & 0.4344  \\
          \hline  
        
    \end{tabular}
    \caption{Comparison of performance of both the models with other methods proposed in the literature on Flickr8k and Flickr30k datasets.}
    \label{comparison_sota}
\end{table}

\begin{table}[h]
    \centering
    \begin{tabular}{p{2.5cm}p{1.5cm}p{1.5cm}p{1.5cm}p{1.5cm}p{1.5cm}p{1.5cm}p{2cm}}
    \hline \noindent
         \begin{tabular}{l} Number  \\ of Layers \end{tabular}
         & BLEU-1
         & BLEU-2
         & BLEU-3
         & BLEU-4
         & METEOR
         & CIDEr
         & ROUGE-L \\ \hline 
          \multicolumn{8}{l}{CNN + CNN Model (Encoder-Decoder framework)} \\ 
          \hline      
         1 & 0.6279 & 0.4439 & 0.3022 & 0.2038 & 0.1924 & 0.4748 & 0.4544\\
         2 & 0.6246 & 0.4350 & 0.2930 & 0.1986 & 0.1915 & 0.4502 & 0.4492 \\
         3 & 0.6193 & 0.4355 & 0.2967 & 0.2009 & 0.1939 & 0.4685 & 0.4507 \\
         4 & 0.6173 & 0.4345 & 0.2964 & 0.2013 & 0.1943 & 0.4670 & 0.4495 \\
         \hline
          \multicolumn{8}{l}{CNN + CNN + Attention Model (Encoder-Decoder framework with Attention)} \\ 
          \hline  
         1 & 0.6257 & 0.4430 & 0.3018 & 0.2040 & 0.1929 & 0.4721 & 0.4541 \\ 
         2 & 0.6291 & 0.4443 & 0.3032 & 0.2060 & 0.1944 & 0.4851 & 0.4556 \\
         3 & 0.6168 & 0.4270 & 0.2894 & 0.1971 & 0.1901 & 0.4447 & 0.4440 \\
         4 & 0.6148 & 0.4255 & 0.2887 & 0.1942 & 0.1908 & 0.4519 & 0.4460  \\ \hline
        
    \end{tabular}
    \caption{Performance of both models with different number of layers on Flickr8k dataset.}
    \label{table_layers_f8k}
\end{table}

\begin{table}[h]
    \centering
    \begin{tabular}{p{2.5cm}p{1.5cm}p{1.5cm}p{1.5cm}p{1.5cm}p{1.5cm}p{1.5cm}p{2cm}}
    \hline \noindent
         \begin{tabular}{l} Number  \\ of Layers \end{tabular}
         & BLEU-1
         & BLEU-2
         & BLEU-3
         & BLEU-4
         & METEOR
         & CIDEr
         & ROUGE-L \\ \hline 
        \multicolumn{8}{l}{CNN + CNN Model (Encoder-Decoder framework)} \\ \hline 
        1 & 0.6432 & 0.4495 & 0.3108 & 0.2123 & 0.1774 & 0.3831 & 0.4344 \\ 
        2 & 0.6393 & 0.4469 & 0.3113 & 0.2154 & 0.1770 & 0.3874 & 0.4346 \\
        3 & 0.6364 & 0.4413 & 0.3051 & 0.2094 & 0.1782 & 0.3838 & 0.4338  \\
        4 & 0.6303 & 0.4363 & 0.3006 & 0.2070 & 0.1763 & 0.3780 & 0.4305  \\ \hline
        \multicolumn{8}{l}{CNN + CNN + Attention Model (Encoder-Decoder framework with Attention)} \\ \hline
        1 & 0.6379 & 0.4445 & 0.3072 & 0.2087 & 0.1770 & 0.3770 & 0.4332 \\
        2 & 0.6393 & 0.4428 & 0.3053 & 0.2088 & 0.1778 & 0.3810 & 0.4344  \\
        3 & 0.6398 & 0.4453 & 0.3077 & 0.2120 & 0.1784 & 0.3923 & 0.4358 \\
        4 & 0.6281 & 0.4345 & 0.2976 & 0.2018 & 0.1754 & 0.3622 & 0.4295  \\
        \hline
    \end{tabular}
    \caption{Performance of both the models with different number of layers on Flickr30k dataset.}
    \label{table_layers_f30k}
\end{table}

\clearpage

\begin{table}[h]
    \centering
    \begin{tabular}{p{2.25cm}|p{3cm}|p{3cm}|p{3cm}|p{3cm}}
    \hline
         Model(Layers)
         &\includegraphics[width=3cm, height=2.75cm]{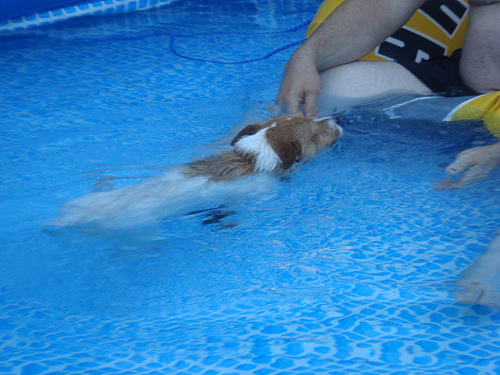}
         &\includegraphics[width=2.75cm, height=3.5cm]{f8klayers/2902269566_419d9f1d8e.jpg}
         &\includegraphics[width=2.75cm, height=3.5cm]{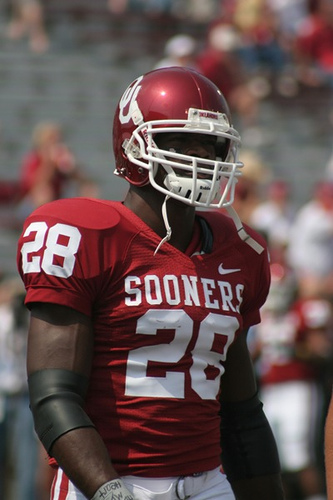}
         &\includegraphics[width=3cm, height=2.75cm]{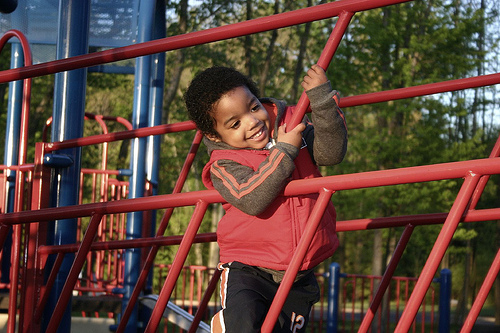}  \\ \hline
         CNN+ CNN(1)
         &a small white dog is jumping into a pool
         &a man riding a bike on a dirt bike
         &a football player in red and white uniform wearing a red and white uniform
         &a little boy in a red shirt is sitting on a swing \\ \hline
         CNN+ CNN(2)
         &a white dog is jumping into a pool
         &a person riding a bike on a dirt bike
         &a football player in a red uniform and red uniform
         &a little girl in a red shirt is sitting on a swing \\ \hline
         CNN+ CNN(3)
         &a small white dog is playing in a pool
         &a person riding a bike on a dirt bike
         &a football player in a red uniform and a red uniform
         &a little boy in a red shirt is jumping on a swing \\ \hline
         CNN+ CNN(4)
         &a white and white dog is playing in a pool
         &a person riding a bike on a dirt bike
         &a football player in a red and white uniform
         &a little girl in a red shirt is sitting on a swing \\ \hline
         CNN+CNN+ Att(1) 
         &a white dog is swimming in a pool
         &a person riding a bike on a dirt bike
         &a football player in a red uniform and a football
         &a little boy in a red shirt is jumping over a swing \\ \hline
         CNN+CNN+ Att(2) 
         &a white dog is jumping over a blue pool
         &a man on a motorcycle rides a dirt bike
         &a football player in a red uniform
         & a little boy in a red shirt is sitting on a swing\\ \hline
         CNN+CNN+ Att(3) 
         &a small white dog is jumping into a pool
         &a person riding a bike on a dirt bike
         &a football player in a red uniform and a red uniform
         &a little boy in a red shirt is jumping over a swing \\ \hline
         CNN+CNN+ Att(4) 
         &a white dog is jumping over a blue pool
         &a man riding a bike on a dirt bike
         &a football player in a red uniform is holding a football
         &a little girl in a pink shirt is sitting on a swing \\ \hline

    \end{tabular}
    \caption{Examples of captions generated by both Encoder Decoder (CNN+CNN) model and Attention based (CNN+CNN+Att) model on Flickr8k dataset. The number of Convolutional layers in the model are in parentheses.}
    \label{table_layers_f8k_examples}
\end{table}

We can observe from the results in Table \ref{comparison_sota} that, on Flickr8k dataset, \textit{CNN+CNN} method has comparable performance with the method proposed in Vinyals \etal \cite{vinyals} and performs better than the method proposed in Karpathy \etal \cite{karpathy} which are encoder-decoder based methods. The method \textit{CNN+CNN+Att}, which is an attention based framework, performs marginally better than \textit{CNN+CNN} method but on BLEU-1 metric it's evaluation score is around 4 points behind the method proposed in Xu \etal \cite{attendtell}. On other metrics the performance is similar for both methods.

On Flickr30k dataset, \textit{CNN+CNN} method performs comparably on most metrics with the method proposed in Vinyals \etal \cite{vinyals} except on BLEU-1 score where it lags behind by around 2 points. However, it should be noted that the results cited in Vinyals \etal\cite{vinyals} were obtained using an ensemble of models whereas the \textit{CNN+CNN} method does not use model ensembling. CNN+CNN method performs better than other encoder-decoder based methods such as the method proposed in Mao \etal \cite{mao}, Donahue \etal \cite{donahue} and Karpathy \etal \cite{karpathy} on most evaluation metrics. In addition, on Flickr30k dataset performance of both \textit{CNN+CNN}
and \textit{CNN+CNN+Att} method is similar to their performance on Flickr8k dataset on most metrics except the CIDEr evaluation metric on which both models perform poorly when trained using Flickr30k dataset.

\begin{table}[]
    \centering
    \begin{tabular}{p{2.25cm}|p{3cm}|p{3cm}|p{3cm}|p{3cm}}
    \hline
         Model(Layers)
         & \includegraphics[width = 3cm, height = 3.5cm]{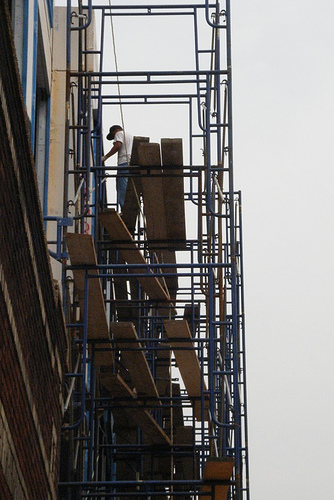}
         & \includegraphics[width = 3cm, height = 2.5cm]{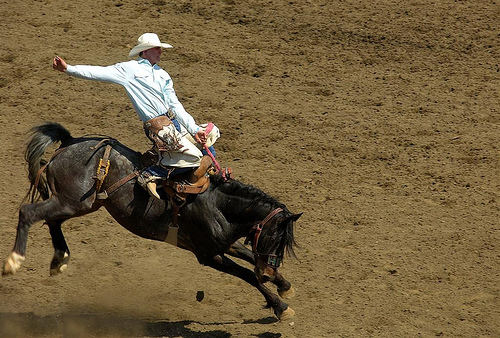}
         & \includegraphics[width = 3cm, height = 2.5cm]{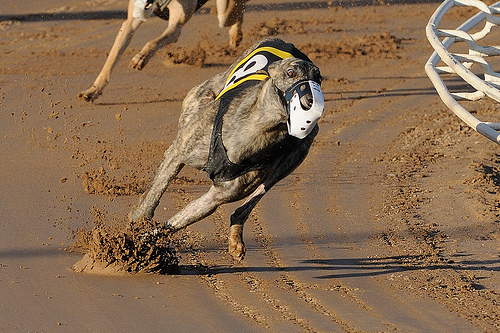}
         & \includegraphics[width = 3cm, height = 2.5cm]{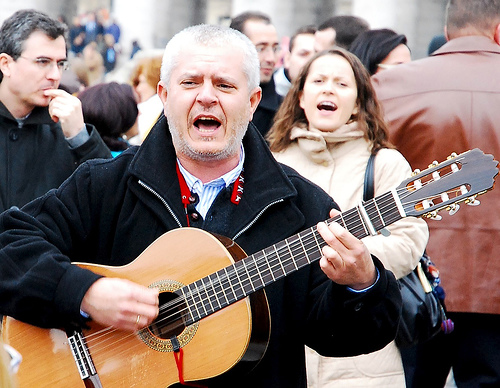} \\ \hline
         CNN+ CNN(1)
         &a construction worker is working on a roof
         &a man in a cowboy hat is riding a horse
         &a brown and white dog is running on a dirt track
         & a man in a black shirt playing a guitar\\ \hline
         CNN+ CNN(2)
         &a construction worker is working on a scaffolding
         &a man is riding a horse in a rodeo
         &two dogs run through the sand
         &a man is playing a guitar and singing \\ \hline
         CNN+ CNN(3)
         &a construction worker is working on a train
         &a man in a cowboy hat is riding a horse
         &two dogs are running in the sand
         &a man is playing a guitar\\ \hline
         CNN+ CNN(4)
         &a construction worker is working on a construction site
         &a man in a cowboy hat is riding a horse
         &two dogs are running in the sand
         &a man in a black shirt is playing a guitar \\ \hline
         CNN+CNN+ Att(1) 
         &a man is working on a ladder
         &a man in a cowboy hat is riding a horse
         &two dogs racing on the beach
         &a man in a black shirt playing a guitar  \\ \hline
         CNN+CNN+ Att(2) 
         &a construction worker is working on a scaffolding
         &a man wearing a cowboy hat is riding a horse
         &two dogs running in the sand
         &a man is playing a guitar \\ \hline
         CNN+CNN+ Att(3) 
         &a man in a white shirt is working on a roof
         &a man is riding a horse
         &a brown and white dog is running in the sand
         &a man in a black shirt plays the guitar \\ \hline
         CNN+CNN+ Att(4) 
         &a man in a white shirt is working on a roof
         &a man is riding a horse
         &two dogs are running in the sand
         & a man is playing a guitar and singing into a microphone \\ \hline

    \end{tabular}
    \caption{Examples of captions generated by both Encoder Decoder (CNN+CNN) model and Attention based (CNN+CNN+Att) model on Flickr30k dataset. The number of Convolutional layers in the model are in parentheses.}
    \label{table_layers_f30k_examples}
\end{table}

\subsection{Depth of Decoder}
\label{decoder_depth}
In Tables \ref{table_layers_f8k} and \ref{table_layers_f30k} we compare the performance of both \textit{CNN+CNN} and \textit{CNN+CNN+Att} methods when trained and evaluated with Flickr8k and Flickr30k datasets, respectively. As we can observe from Tables \ref{table_layers_f8k} and \ref{table_layers_f30k}, on both Flickr8k and Flickr30k datasets, \textit{CNN+CNN} method achieves best performance when one Convolutional layer is used and for \textit{CNN+CNN+Att} method, the best performance is achieved using two stacked Convolutional layers. Using more layers does not offer improvement in performance. 

In Tables \ref{table_layers_f8k_examples} and \ref{table_layers_f30k_examples}, we present some examples of captions generated using both CNN+CNN and CNN+CNN+Att methods trained using different number of layers on Flickr8k and Flickr30k datasets, respectively.

\subsection{Image Transforms as Data Augmentation}
\label{image_data_augmentation}
Here, we describe the evaluation scores of both \textit{CNN+CNN} and \textit{CNN+CNN+Att} methods using five image data augmentation techniques as described in Section \ref{data_augmentation}. We use Flickr8k dataset for training and evaluation. We can observe from evaluation scores in Table \ref{table_transforms} that most data augmentation techniques don't offer tangible gains in performance.
Use of Random Horizontal transform shows some improvement in evaluation scores for both the methods whereas both Random Vertical transformation, where the image is flipped upside down and Random Rotate transform, where the image is rotated by 90, 180 or 270 degrees, show decrease in performance for both methods with respect to most evaluation metrics. Similarly, Perspective Transformation also does not offer improvement in Captioning Performance. In Table \ref{table_transforms_results}, we present a few examples of captions generated with both \textit{CNN+CNN} and \textit{CNN+CNN+Att} methods when trained with different image transform methods for data augmentation. Most methods generate similar sentences and there does not appear much variation in the quality of captions generated.

\begin{table}[]
    \centering
    \begin{tabular}{p{3cm}p{1.5cm}p{1.5cm}p{1.5cm}p{1.5cm}p{1.5cm}p{1.5cm}p{2cm}}
    \hline \noindent
         Image Transform
         & BLEU-1
         & BLEU-2
         & BLEU-3
         & BLEU-4
         & METEOR
         & CIDEr
         & ROUGE-L \\ \hline 
          \multicolumn{8}{l}{CNN + CNN method} \\ 
          \hline      
          No transformation & 0.6279 & 0.4439 & 0.3022 & 0.2038 & 0.1924 & 0.4748 & 0.4544\\
          Random Horizontal & 0.6275 & 0.4485 & 0.3063 & 0.2053 & 0.1953 & 0.4779 & 0.4561 \\
          Random Vertical & 0.6222 & 0.4341 & 0.2926 & 0.1948 & 0.1890 & 0.4427 & 0.4476 \\ 
          Random Flip & 0.6281 & 0.4462 & 0.3045 & 0.2057 & 0.1921 & 0.4698 & 0.4531 \\
          Random Rotate & 0.6115 & 0.4229 & 0.2831 & 0.1885 & 0.1845 & 0.4205 & 0.4379 \\
          Random Perspective & 0.6253 & 0.4428 & 0.3002 & 0.2004 & 0.1910 & 0.4562 & 0.4506  \\
          \hline
        \multicolumn{8}{l}{CNN+CNN+Att method} \\ \hline 
        No transformation & 0.6291 & 0.4443 & 0.3032 & 0.2060 & 0.1944 & 0.4851 & 0.4556 \\
        Random Horizontal & 0.6324 & 0.4509 & 0.3094 & 0.2099 & 0.1928 & 0.4798 & 0.4572 \\
        Random Vertical & 0.6162 & 0.4303 & 0.2912 & 0.1939 & 0.1863 & 0.4294 & 0.4441  \\
        Random Flip & 0.6234 & 0.4386 & 0.2974 & 0.1984 & 0.1891 & 0.4523 & 0.4500 \\
        Random Rotate & 0.6077 & 0.4192 & 0.2797 & 0.1868 & 0.1824 & 0.4097 & 0.4336 \\
        Random Perspective & 0.6256 & 0.4414 & 0.3023 & 0.2057 & 0.1909 & 0.4605 & 0.4483 \\ \hline
    \end{tabular}
    \caption{Performance of Model with different Image Transforms on Flickr8k dataset.}
    \label{table_transforms}
\end{table}

\begin{table}[]
    \centering
    \begin{tabular}{p{2.25cm}|p{3cm}|p{3cm}|p{3cm}|p{3cm}}
    \hline
         Transforms used 
         & \includegraphics[width=3cm, height = 2.5cm]{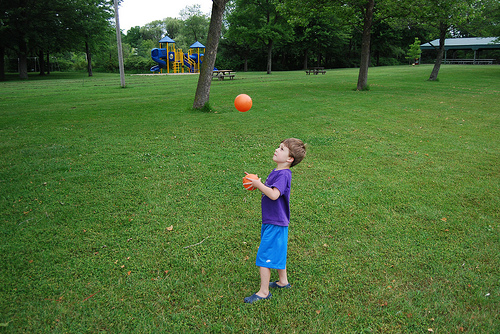}
         & \includegraphics[width=3cm, height = 2.5cm]{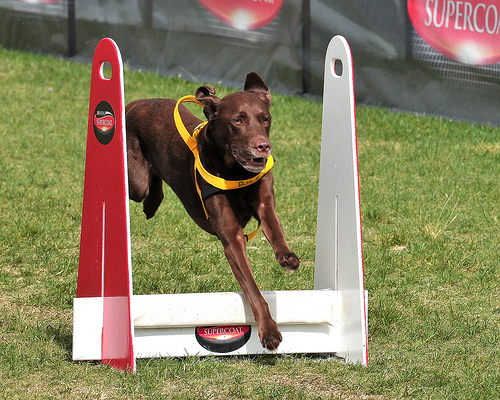}
         & \includegraphics[width=3cm, height = 3.5cm]{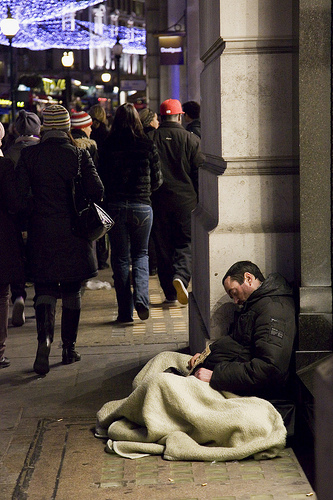}
         & \includegraphics[width=3cm, height = 2.5cm]{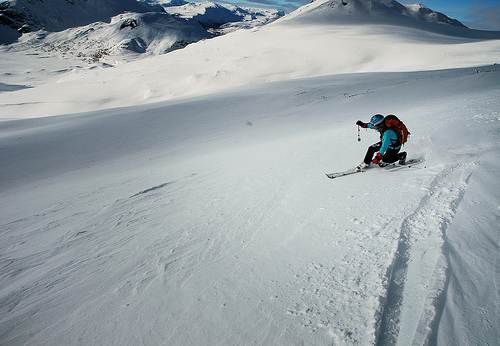} \\ \hline
         &(a)&(b)&(c)&(d) \\ \hline
         \multicolumn{5}{l}{\hspace{3cm} Method used: \textit{CNN+CNN}} \\ \hline 
         Random Horizontal 
         & a boy in a red shirt is running on the grass
         & a black dog is running through the grass
         & a group of people are standing in front of a building
         & a skier skiing down a snowy hill \\
         Random Vertical
         & a little boy in a blue shirt is playing with a soccer ball
         &a black dog is running through the grass
         &a group of people stand in front of a building
         & a skier is skiing down a snowy hill \\
         Random Flip
         &a little boy in a blue shirt is running through the grass
         &a black dog is running through the grass
         &a group of people stand in front of a building
         &a skier skiing down a snowy hill \\
         Random Rotate
         & a little boy in a blue shirt is running through a field
         &a black dog is running through the grass
         &a group of people are standing on a street
         & a skier is skiing down a snowy hill\\
         Random Perspective
         & a boy in a blue shirt is playing with a ball
         &a black dog is running through the grass
         &a group of people stand in front of a building
         & a skier skiing down a snowy hill\\ \hline
         \multicolumn{5}{l}{\hspace{3cm} Method used: \textit{CNN+CNN+Att}} \\ \hline 
          Random Horizontal 
         &a boy in a red shirt is running on the grass
         &a black dog is running through the grass
         &a group of people are standing in front of a building
         & a skier skiing down a snowy hill\\
         Random Vertical
         &a little boy in a blue shirt is running through a field
         &a black dog is running through the grass
         &a group of people stand in front of a building
         & a skier in the snow\\
         Random Flip
         &a boy in a blue shirt is playing with a soccer ball
         &a black dog is running through the grass
         &a group of people are standing outside
         & a skier is skiing down a snowy hill\\
         Random Rotate
         &a boy in a blue shirt is running on the grass
         &a black dog is running through the grass
         &a group of people are standing on a street
         &a skier is skiing down a snowy hill \\
         Random Perspective
         &a boy in a red shirt is running on the grass
         &a black dog is running through the grass
         &a group of people stand in front of a building
         &a skier skiing down a snowy hill \\ \hline
    \end{tabular}
    \caption{Examples of captions generated by the Encoder-Decoder (\textit{CNN+CNN}) and Attention based (\textit{CNN+CNN+Att}) methods when trained with various Image Transformations for Data Augmentation using Flickr8k dataset.}
    \label{table_transforms_results}
\end{table}

\begin{table}[]
    \centering
    \begin{tabular}{p{2.25cm}p{1.5cm}p{1.5cm}p{1.5cm}p{1.5cm}p{1.5cm}p{1.5cm}p{2cm}}
    \hline \noindent
         Max. Sentence Length
         & BLEU-1
         & BLEU-2
         & BLEU-3
         & BLEU-4
         & METEOR
         & CIDEr
         & ROUGE-L \\ \hline 
         \multicolumn{8}{l}{\textit{CNN+CNN}} \\ \hline 
         10& 0.6275 & 0.4395 & 0.2985 & 0.2047 & 0.1682 & 0.3714 & 0.4291 \\
         15& 0.6246 & 0.4350 & 0.2930 & 0.1986 & 0.1915 & 0.4502 & 0.4492 \\
         20& 0.6111 & 0.4310 & 0.2928 & 0.1965 & 0.1917 & 0.4601 & 0.4506   \\
         25& 0.6102 & 0.4343 & 0.3005 & 0.2062 & 0.1948 & 0.4763 & 0.4546 \\
         30& 0.5651 & 0.3996 & 0.2713 & 0.1820 & 0.1894 & 0.4559 & 0.4448  \\
         35& 0.5576 & 0.3928 & 0.2677 & 0.1787 & 0.1894 & 0.4569 & 0.4439  \\
         40& 0.5327 & 0.3748 & 0.2545 & 0.1698 & 0.1891 & 0.4553 &  0.4411 \\
         \hline
         \multicolumn{8}{l}{\textit{CNN+LSTM}} \\ \hline
         10 & 0.4541 & 0.2953 & 0.1916 & 0.1253 & 0.1708 & 0.6253 & 0.3943 \\
         15 & 0.5898 & 0.4128 & 0.2825 & 0.1923 & 0.1966 & 0.5549 & 0.4490 \\
         20 & 0.6126 & 0.4317 & 0.3000 & 0.2065 & 0.2023 & 0.5326 & 0.4599 \\
         25 & 0.6252 & 0.4402 & 0.3042 & 0.2091 & 0.2000 & 0.5419 & 0.4610 \\
         30 & 0.6236 & 0.4394 & 0.3044 & 0.2076 & 0.2014 & 0.5205 & 0.4591 \\
         35 & 0.6230 & 0.4401 & 0.3066 & 0.2121 & 0.1978 & 0.5281 & 0.4594 \\
         40 & 0.6190 & 0.4383 & 0.3062 & 0.2122 & 0.2030 & 0.5497 & 0.4628 \\ \hline
         
    \end{tabular}
    \caption{Performance of \textit{CNN+CNN} and \textit{CNN+LSTM} methods on Flickr8k dataset when trained with different maximum sentence length values.}
    \label{table_sentence_length}
\end{table}

\subsection{Variation in Sentence length}
\label{sentence_length}
Here, we experiment with different values of maximum sentence length used for selecting sentences during training. We have presented results of both \textit{CNN+CNN} and \textit{CNN+LSTM} methods when trained using different values of maximum length of sentences used for training.  As we can observe from results in Table \ref{table_sentence_length}, including sentences of length higher than 20 words leads to rapid decrease in performance scores for \textit{CNN+CNN} method with respect of most evaluation metrics. In addition, there is considerable variation in the distribution of scores obtained from different metrics. At lower values of maximum sentence length such as values of 10 and 15, better BLEU scores are obtained whereas CIDEr scores are considerably lower than those obtained with higher values of maximum sentence length such as 25 and above. On comparison with evaluation scores for \textit{CNN+LSTM} method, we find that using LSTM decoder leads to significant improvement in CIDEr scores. In addition, when using LSTM decoder, at higher values of maximum sentence length the evaluation scores are considerably better than those generated when model is trained with lower values of maximum sentence length.

We speculate that this may be due to the fact than an LSTM decoder is much more efficient at modelling dependencies between words which occur farther apart in the sentence.

In Table \ref{sentence_length_examples} we present a few examples of captions generated when the \textit{CNN+CNN} model is trained using different values of maximum sentence length. When the model is trained with longer sentences, in some cases the generated sentence is composed of incoherent and multiple repetitions of phrases. Also model trained with longer sentences usually generates more number of \textit{UNK} token which is the placeholder word for tokens which are not present in the vocabulary.

\begin{table}[h]
    \centering
    \begin{tabular}{p{1.5cm}|p{3.15cm}|p{3.15cm}|p{3.15cm}|p{3.15cm}}
    \hline
         Maximum Sentence Length
         & \includegraphics[width=3cm, height = 2.5cm]{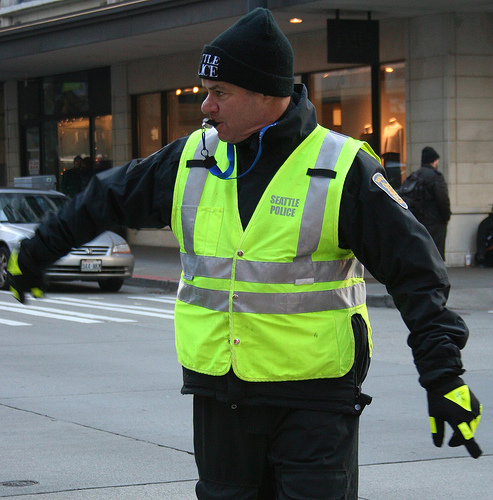}
         & \includegraphics[width=3cm, height = 3.5cm]{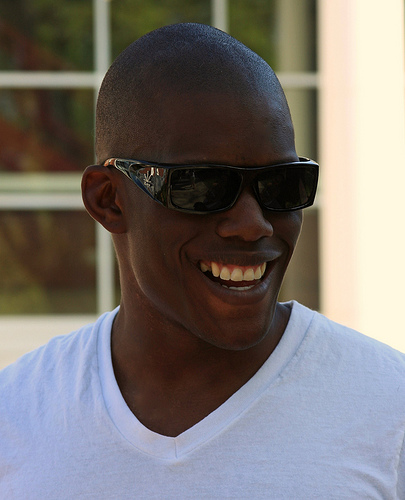}
         & \includegraphics[width=3cm, height = 2.5cm]{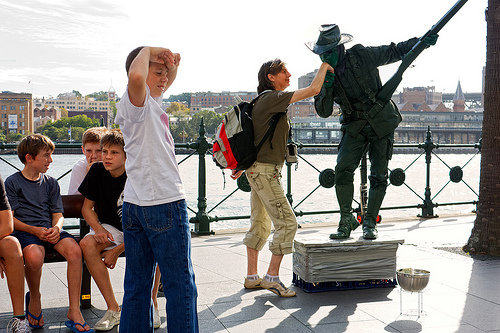}
         & \includegraphics[width=3cm, height = 2.5cm]{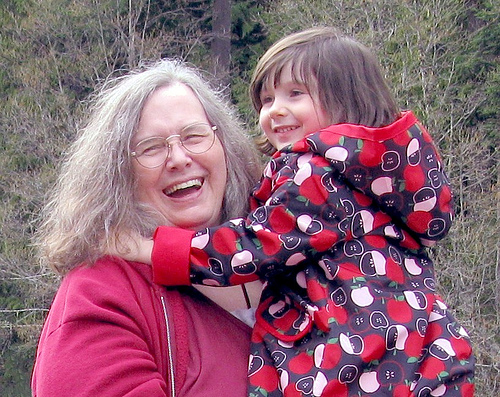} \\ \hline
         10
         & a man wearing a black jacket and black
         & a man wearing a black shirt and
         & a group of people are standing on a
         & a girl in a red shirt and
         \\ \hline
         15
         & a man in a yellow shirt is standing on a street
         & a man in a blue shirt and sunglasses
         & a group of people are standing on a sidewalk
         & a little girl in a red shirt and a girl in a red shirt
         \\ \hline
         20
         & a man wearing a black jacket and a woman in a black jacket and a woman in a black
         & a man wearing a black shirt and sunglasses is wearing a black shirt and sunglasses is wearing a black
         & a group of people are standing on a sidewalk
         & a little girl in a red shirt and a girl in a red shirt and a girl in a
         \\ \hline
         25
         & a man in a yellow jacket is standing on a street
         & a man wearing a blue shirt and sunglasses is standing in front of a UNK
         & a man in a black shirt and a woman in a black shirt and a woman in a black shirt and a woman in
         & a little girl in a red shirt and a girl in a red shirt and a girl in a red shirt and a girl \\ \hline
         30
         & a man in a yellow jacket and a woman in a yellow jacket and a woman in a yellow jacket and a woman in a yellow jacket and a
         & a man wearing a black shirt and sunglasses
         & a group of people are standing on the sidewalk
         & a little girl in a red shirt and a girl in a red shirt and a girl in a red shirt and a girl in a red shirt and \\ \hline
         35
         & a man in a yellow shirt and a woman in a yellow jacket is standing on a street
         & a man in a black shirt and sunglasses is wearing a blue shirt and sunglasses is standing in front of a crowd
         & a group of people are standing on a sidewalk
         & a little girl in a red shirt and a girl in a red shirt is holding a picture
         \\ \hline
         40
         & a man in a yellow jacket is standing in the street
         & a man in a blue shirt and sunglasses is standing in front of a UNK
         & a man in a white shirt and a woman in a black shirt and a woman in a black shirt and a woman in a black shirt and a woman in a black shirt and a woman in a
         & a little girl in a red shirt is holding a UNK   \\ \hline
    \end{tabular}
    \caption{Examples of captions generated by the Encoder-Decoder method with Convolutional Decoder (\textit{CNN+CNN}) when trained using different values of maximum lengths of sentences.}
    \label{sentence_length_examples}
\end{table}

\section{Conclusion}
\label{conclusion}
In this work, we perform extensive experiments to analyse various aspects of Encoder-Decoder and Attention based Image Captioning using Convolutional Decoder. We analyse network depth in terms of number of stacked Convolutional layers, Image Data Augmentation techniques such as Flipping, Rotation and Perspective Transformation of images and evaluated the model after training with different values of maximum sentence length. We have observed that best scores are obtained using a single layered Convolutional Decoder for encoder-decoder framework and two layered Convolutional Decoder for Attention based framework. In addition, image transforms used as Data Augmentation techniques do not provide performance improvements except for Random Horizontal transforms which provide slight improvement in evaluation scores. Furthermore, Convolutional Decoders provide good performance when trained with sentences of length lower than 15 sentences and suffer from drastic reduction in performance when trained using sentences of longer length.

\clearpage
\subsubsection*{Acknowledgments.}
We would like to thank the Language Processing Laboratory and Medical Imaging Laboratory for providing GPU equipped workstations which were indispensable for this work.





\end{document}